\documentclass[runningheads]{llncs}
\usepackage[T1]{fontenc}
\usepackage{graphicx}
\usepackage{multirow,amsmath,algorithmic,algorithm,amsfonts,booktabs}
\begin{document}
\title{Task Consistent Prototype Learning for Incremental Few-shot Semantic Segmentation}
\titlerunning{Task Consistent iFSS}
%
\author{Wenbo Xu\inst{1} \and
Yanan Wu\inst{2} \and Haoran Jiang\inst{1} \and Yang Wang\inst{3} \and Qiang Wu\inst{1} \and Jian Zhang\inst{1}}
\authorrunning{X. Wenbo et al.}
%
\institute{Faculty of Engineering and Information Technology, University of Technology Sydney, Sydney, 2007, Australia\\
\email{Wenbo.xu@student.uts.edu.au}\\
\and
School of Computer and Information Technology, Beijing Jiaotong University, Beijing, 100044, China \\
\and Department of Computer Science and Software Engineering, Concordia University, Montreal, H3G2J1, Canada
}
\maketitle              
\begin{abstract}
Incremental Few-Shot Semantic Segmentation (iFSS) tackles a task that requires a model to continually expand its segmentation capability on novel classes using only a few annotated examples. Typical incremental approaches encounter a challenge that the objective of the base training phase (fitting base classes with sufficient instances) does not align with the incremental learning phase (rapidly adapting to new classes with less forgetting). This disconnect can result in suboptimal performance in the incremental setting. This study introduces a meta-learning-based prototype approach that encourages the model to learn how to adapt quickly while preserving previous knowledge. Concretely, we mimic the incremental evaluation protocol during the base training session by sampling a sequence of pseudo-incremental tasks. Each task in the simulated sequence is trained using a meta-objective to enable rapid adaptation without forgetting. To enhance discrimination among class prototypes, we introduce prototype space redistribution learning, which dynamically updates class prototypes to establish optimal inter-prototype boundaries within the prototype space. Extensive experiments on iFSS datasets built upon PASCAL and COCO benchmarks show the advanced performance of the proposed approach, offering valuable insights for addressing iFSS challenges.
\end{abstract}
\section{Introduction}
Deep learning models have made remarkable strides in semantic segmentation tasks by training on extensive datasets with rich annotations. In an effort to alleviate the burden of data annotation, Few-shot Semantic Segmentation (FSS) ~\cite{BAM,xu2023masked} has been introduced, aiming to rapidly adapt to novel classes with minimal labeled data rapidly. However, FSS frameworks typically operate under a fixed output space assumption, where the number of target classes is predetermined. This limitation constrains the practicality and scalability of deployment in real-world scenarios where the total number of categories is uncertain, and new class objects may emerge over time.

In this work, we address a more challenging and practical scenario where the model continuously encounters a stream of new image data containing instances of previously unseen classes. The objective is to update a model to effectively segment new classes using a few annotated samples while retaining its segmentation capability on existing seen classes. The task known as Incremental Few-Shot Semantic Segmentation (iFSS) in the existing literature~\cite{protoiFSS,advancing}, is inspired by few-shot class incremental learning (FSCIL)~\cite{fscil}. It shares two common challenges, namely catastrophic forgetting of learned knowledge and overfitting to a limited number of novel class examples. This arises due to the absence of access to previous session data during the incremental learning sessions. When updating parameters with imbalanced novel class data (where the number of novel classes is considerably smaller compared to base classes), the model tends to exhibit a strong bias towards novel classes in pursuit of rapid adaptation. Consequently, there is a risk of aggressively overwriting crucial knowledge related to old classes in an attempt to accommodate the latest instances, resulting in a loss of generalization ability.

The challenges mentioned above stem from the task misalignment inherent in existing iFSS methods~\cite{protoiFSS,advancing}. These methods begin by initializing a model that effectively predicts the base classes through classical supervised learning during the base training session. However, in subsequent incremental sessions, the focus shifts to pursuing fast adaptation to novel classes with less forgetting. To overcome this drawback, we propose a meta-learning~\cite{mefscil,metagcd} based approach that directly learns to incrementally adapt to novel classes conditioned on a few examples. This is achieved by simulating the incremental few-shot scenario during base session training. Concretely, we sample a sequence of pseudo incremental tasks from the base class dataset. For each pseudo task, the model performs fast adaptation with a few new class examples and updates itself. Then the meta loss is calculated by measuring the performance of the updated model on the test images of both the old and the new classes. The object of the meta loss is meta loss is to incentivize the model to incrementally learn new classes while minimizing the forgetting of the old ones.

Recently, some FSCIL studies~\cite{fewincreEvoled,fewincreself} trains a backbone network on the base session and subsequently keep its parameters fixed during incremental sessions to maintain a consistent feature extractor. However, in these methods, the feature extractor remains static, implying that the feature space distributed for the base class is reused to accommodate additional classes. Our approach relies on prototype learning, wherein a prototype for a novel class is constructed from its features, forming a prototype classifier alongside the prototypes of the base classes. When generating a prototype for a new class, it may be positioned close to the prototypes of the base classes in the feature space. This can result in interference, where a pixel will produce high similarity scores with both new and old prototypes, leading to catastrophic forgetting. 

To optimize the prototype generation process, we propose a Prototype Space Re-distribution Learning (PSRL) to incrementally learn novel class prototypes and adaptively allocate base and novel prototypes into a latent prototype space, maintaining optimal prototype boundaries. Specifically, we fix the pre-trained feature backbone to preserve a unified feature extractor and introduce a prototype projector mapping intermediate class vectors to a subspace for dynamic prototype distribution. The redistribution process aims to enhance discrimination between new and old class prototypes, thereby improving segmentation performance. Furthermore, it regulates the updated base prototypes placed near their previous position to prevent prototype misalignment, effectively mitigating knowledge forgetting. The contributions of this work are summarized as:

\begin{itemize}

    \item We introduce a meta-learning approach that closely aligns the base learning objective with the incremental evaluation protocol. Through training with a series of pseudo incremental tasks, our method directly optimizes the model to enhance the discovery of novel objects while mitigating forgetting

    \item We present Prototype Space Re-distribution Learning (PSRL), a method that incrementally learns novel classes while considering inter-prototype discrimination and maintaining base prototype consistency. This approach alleviates catastrophic forgetting of base classes and facilitates rapid adaptation to novel classes.

    \item Extensive experiments on dedicated iFSS benchmark from PASCAL and COCO datasets demonstrate the proposed method outperforms several counterparts.
\end{itemize}
\begin{figure*}[t]
	\centering
	\includegraphics[width=0.8\textwidth]{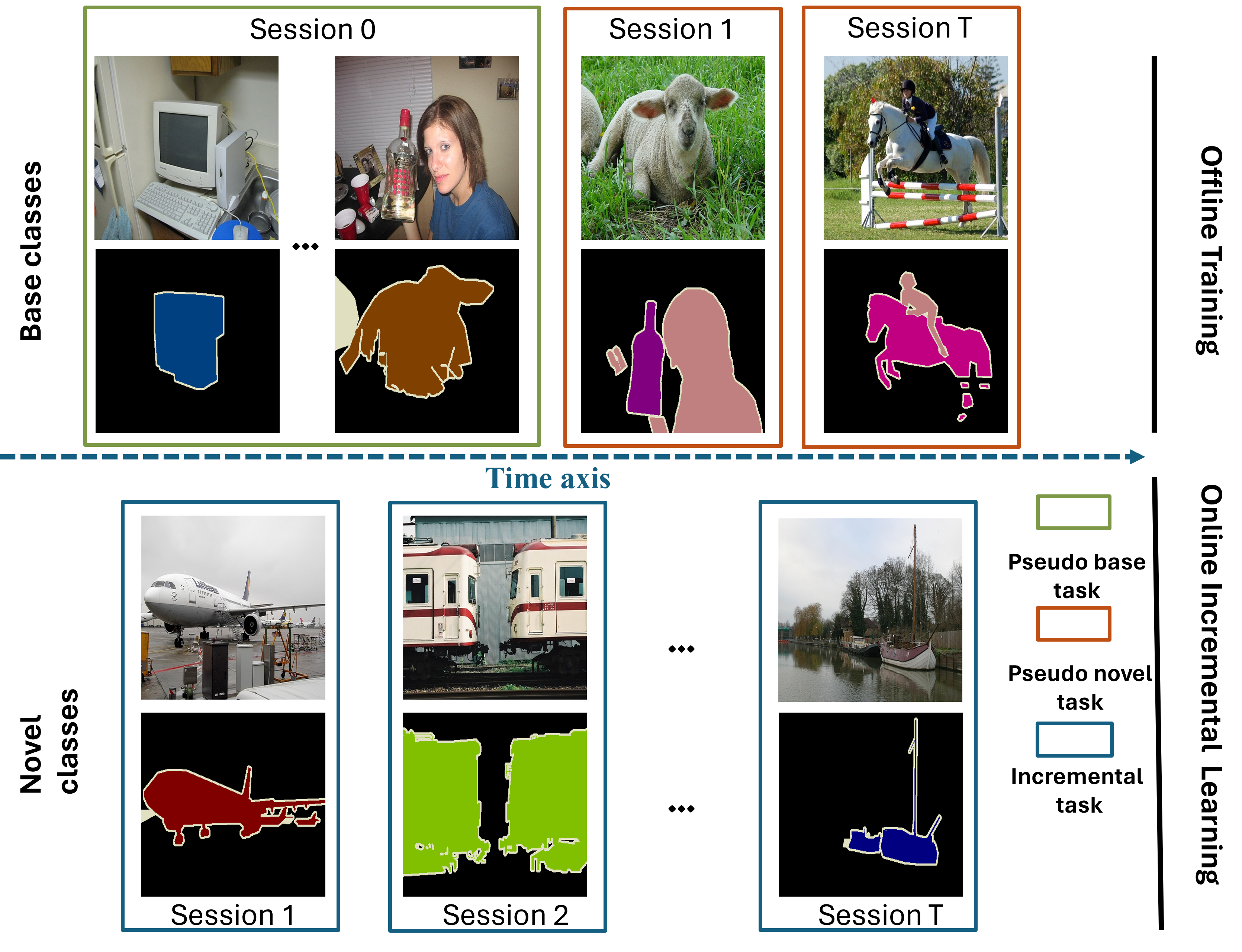}
	\caption{Illustration of the evaluation protocol and our meta-training process. During the online incremental learning stage, the model undergoes training solely on new classes within each incremental session, while evaluation is conducted on all classes encountered thus far. Our strategy aims to replicate this evaluation protocol during the offline base class training stage. This is accomplished by randomly sampling a large portion of base class images to constitute the pseudo base dataset, with the remaining classes forming the pseudo novel classes. Initially, the model trains on the pseudo base dataset and subsequently adapts to the pseudo novel classes. This approach enables the model to learn how to swiftly identify new classes while retaining the ability to segment previously encountered ones}
	\label{fig:iFSS}
\end{figure*}

\section{Related work}
\subsection{Semantic Segmentation}
Semantic segmentation has witnessed significant advancements through the development of deep learning models. Fully Convolutional Networks (FCNs)~\cite{FCN} marked a significant milestone, enabling end-to-end learning for pixel-wise classification. Building upon this foundation, numerous Convolutional Neural Networks (CNNs) based architectures have been designed in aspects of optimal encoder-decoder frameworks~\cite{deeplab}, pyramid pooling~\cite{pspnet} and multi-scale feature fusion~\cite{FPN}. While CNNs progressively build context through layers, transformers inherently consider the entire image at each stage, allowing them to capture long-range dependencies more effectively. This has led to the development of transformer-based models that introduce strong feature representation~\cite{pyramidTF}, hybrid CNN-Transformer architectures~\cite{segformer}, and cross-attention decoders~\cite{segmenter,maskFormer}. Despite their impressive performance, these models typically require a substantial amount of mask-annotated data for training and are limited to predefined categories.
\subsection{Few Shot Semantic Segmentation}
To reduce the expenses associated with annotating segmentation data, researchers introduced few-shot semantic image segmentation (FSS). This approach aims to accurately segment objects in an image using only a small number of labeled examples per class. Drawing inspiration from few-shot learning~\cite{prototypical}, FSS models employ a two-branch architecture where a support branch learns class-wise prototypes from a small set of labeled images (support images), and the query image is segmented by comparing each pixel to the support class prototypes. Recent advancements in FSS mainly design models from the aspects of generating versatile prototypes and learning reliable feature correspondence. The prototype optimization strategy~\cite{adaptiveProto,xu2023masked} aims at compressing abstract class information into one or multiple prototypes that enable the model to perform effective feature guidance. The latter encourages the model to consider the most related information between the query-support images during segmentation by learning dense feature correspondence~\cite{HDMNet,HSNet}. 
Despite the progress made in few-shot semantic segmentation (FSS) methods, they specialize primarily in identifying a single novel class by generating a binary foreground-background mask. In contrast, our prototype-based model addresses a more demanding and practical scenario where the model must segment all classes it has seen thus far.

\subsection{Incremental Learning}
Incremental learning (IL), also known as lifelong or continuous learning, is an approach within machine learning that focuses on the model's ability to learn continuously, accommodating new knowledge while retaining previously learned information. IL methods can be broadly classified into replay-based~\cite{directrelay2} and regularization-based~\cite{zhu2022feature}. In replay-based methods, samples of previous tasks are either stored or generated at first and then replayed when learning the new task. Zhu et al.~\cite{zhu2022feature} propose to store the same number of old samples as each new class to form a joint set during its incremental learning process. Regularization-based methods protect old knowledge from being covered by imposing constraints on new tasks. In iFSS, Cermelli et al.~\cite{protoiFSS} introduced a prototype-based distillation loss to force the current model to retain scores for old classes, thereby preventing forgetting. Guangchen et al.~\cite{iFSSEmbed} proposes an embedding adaptive-update strategy to prevent catastrophic forgetting, where hyper-class embeddings remain fixed to preserve existing knowledge. To mitigate feature embedding bias, Kai et al.~\cite{GFSSopen} presents class-agnostic foreground perception across multiple targets. Different from those methods, our method exploits the prototype classifier to remember knowledge and directly optimize the learning process with meta-learning tasks.

\section{Method}
\subsection{Problem setting}
iFSS addresses the challenge of updating a pre-trained segmentation model to accommodate newly introduced classes over time, utilizing limited annotated examples for each novel class. Specifically, let $\mathcal{D}^t_{train/test} = \{ \mathcal{I}^t_n, \mathcal{M}^t_n\}$, $n \in \{1,2, \ldots, K\}, t \in \{1,2,...,T\}$, denote a sequence of the training and testing sets of image $\mathcal{I}_{train/test}^t$ and their corresponding semantic label masks $\mathcal{M}_{train/test}^t$. The label classes $\mathcal{C}^t$ of each set are disjoint, such that $\mathcal{C}^i \cap \mathcal{C}^j = \emptyset, \forall i \neq j$. iFSS comprises a base session with abundant labeled training images from $D_{train}^0$ and a sequence of incremental sessions with only a few training images for each novel class from $\{D_{train}^1, D_{train}^2,..., D_{train}^T\}$. We undertake offline training in the base session to initialize a model using base classes $\mathcal{C}^0$. After the base session, the model is expected to adapt to new classes $\mathcal{C}^t (t > 0)$ with a few examples in the subsequent incremental sessions. Note that at the $t^{th}$ session, the model has access only to $D_{train}^t$ for training and then is evaluated on test images containing all the encountered classes so far, i.e. $\{D_{test}^0 \cup D_{test}^1 ... \cup D_{test}^t\}$
\vspace{-1.0em}

\begin{figure*}[t]
    \centering
    \includegraphics[width=1.0\textwidth]{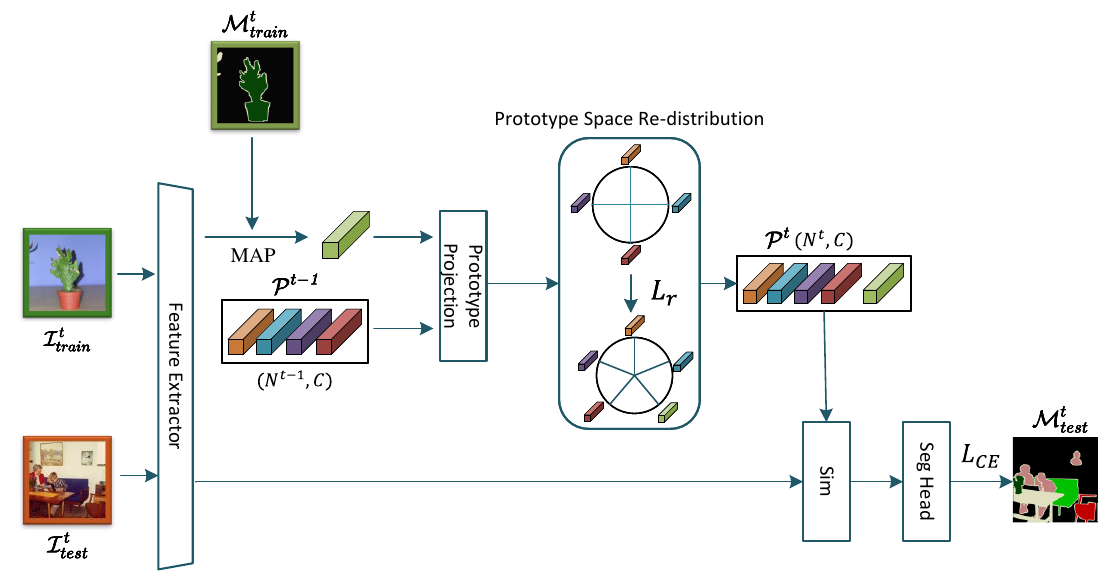}
   \caption{The proposed prototype-based approach utilizes masked average pooling (MAP) to derive the novel class prototype. Subsequently, all prototypes are projected into a latent prototype space for redistribution. The resulting prototypes form a new classifier $\mathcal{P}^t$ capable of identifying both base and novel classes. This process is considered as a sequential task of the meta-learning optimization. In the online incremental sessions, the feature extractor remains frozen, and only the prototype projector and segmentation head are updated}
    \label{fig5.1}
\end{figure*}
\subsection{Prototype Space Re-distribution Learning}

\textbf{Prototype-based Semantic Segmentation}. As introduced in \cite{GFSS}, typical prototype-based few-shot semantic segmentation frameworks comprise a feature extractor and a prototype classifier. Feature extractor transforms the input image $\mathcal{I} \in \mathbb{R}^{h \times w \times 3}$ into a feature embedding $\mathcal{F} \in \mathbb{R}^{w \times h \times d}$ in a latent space. Subsequently, a prototype classifier $\mathcal{P} \in \mathcal{R}^{N \times d}$ is trained to perform pixel-wise predictions for $N$ classes on $\mathcal{F}$. For iFSS, our objective is to progressively expand the base prototype classifier $\mathcal{P}^0$ with prototypes of novel classes, facilitating the continuous segmentation of newly encountered classes without forgetting prior knowledge. Formally, in an N-class K-shot incremental session (N novel classes and each novel class has K training samples), all training samples $\mathcal{I}_{c,n}^t$ are first processed by a feature extractor $f$ and mask average pooling. Subsequently, these samples are averaged over $K$ shots to create $N$ prototypes, denoted as $p^t_c(c \in\{1,2, \ldots, N\})$.
\begin{equation}
p^t_c=\frac{1}{K} \sum_{n=1}^K \frac{\sum_{h, w}\left[ \mathcal{M} _{c,n}^t \circ f \left( \mathcal{I}^t_{c,n} \right) \right]_{h, w}}{\sum_{h, w}\left[\mathcal{M}_{c,n}^t\right]_{h, w}}
\label{eq1},
\end{equation}
where $\mathcal{I}_{c,n}^t$ denotes the $n$-th training image of class $\mathbf{c}$. $\mathcal{M}_{c,n}^t \in \mathbb{R}^{h, w, 1}$ is the class mask for class $\mathbf{c}$ on feature $f\left( \mathcal{I}^t_{c,n} \right) \in \mathbb{R}^{h, w, d}$.  After obtaining $N$ prototypes, the prediction of pixel $i$ of $\mathcal{F}$ is assigned according to the normalized cosine similarity score $S_{i,c}(\mathcal{F})$ between features and the class prototype $p^t_c$ as:
\begin{equation}
S_{i,c}(\mathcal{F})=\frac{\exp \left( Sim(\mathcal{F}_i, \mathbf{p^t_c}) / \tau\right)}{\sum_{j=1}^N \exp \left(Sim(\mathcal{F}_i, \mathbf{p^t_j}) / \tau\right)},
\end{equation}
where $\mathcal{F}_i \in \mathbb{R}^d$ are the positional features extracted from input image $\mathcal{I}$, $N$ represents the cumulative category of prototype vectors up to session $t$, and $\tau$ is a temperature parameter that controls the concentration level of the distribution~\cite{tua}. $Sim(,) = \frac{{\mathcal{F}_i^{\top}}  \mathbf{p^t}}{\|\mathcal{F}_i\|\| \mathbf{p^t} \|}$ is the cosine similarity metric that measures the pixel classification score.

Training with few-shot examples in novel class sessions inevitably leads to overfitting and has the potential to undermine the feature extraction capabilities of the pre-train backbone network. Given that our prototype classifier encompasses both base and newly encountered classes, and we have no access to base examples during incremental learning, modulating the extractor may map new classes into a disparate feature space from that of base classes. Therefore, to ensure consistent feature mapping, the backbone is consistently maintained in a fixed state. However, the newly added prototypes may lie close to the base-class prototypes because the prototype is derived from a fixed feature space that is tailored for base classes. To discriminate novel prototypes from their base counterparts, we introduce the prototype projector $\mathbf{g}$ to map the current prototypes into a latent prototype space where base and novel prototypes are adaptively distributed to achieve two objectives: i) ensuring clear inter-prototype discrimination among base and novel prototypes for fast adaptation to new classes, and ii) minimizing the displacement of base prototypes away from their original positions to prevent catastrophic forgetting and maintain alignment between features and prototypes. Accordingly, we propose a novel prototype redistribution loss that places the new class prototype $p^t_i$ at a position far from base prototypes $P^{t-1}_j$ and relocates base classes to a near-optimal position as:
\begin{equation}
\mathcal{L}_r=\frac{\sum_{i=1}^{N^b}\sum_{j=1}^{N^t} Sim\left(P^{t-1}_i, P^t_j\right)}{\sum_{i=1}^{N^b} Sim(P^{t-1}_i,\hat{P}^{t-1}_i)},
\end{equation}
where $N^b, N^t$ are the class prototype number of previous sessions $[0,1,...,t-1]$ and current session $t$. $\hat{P}^{t-1}_i$ represents the redistributed prototype vector derived from the base prototype $P^{t-1}_i$. We utilize cosine distance as the metric for the similarity matrix. The loss function $\mathcal{L}_r$ is designed to minimize the similarity between new class prototypes and base prototypes, concurrently maximizing the similarity between the original base prototypes and their respective redistributions.

\subsection{Learning to Incrementally learn}
The core idea underlying our approach is meta-learning inspired by MAML~\cite{maml} for few-shot tasks. During the meta-training phase, the model is trained with a set of novel class adaptation tasks that are formulated as few-shot learning problems, aiming to simulate the scenario encountered during meta-testing. In iFSS, the online incremental stage closely resembles the "meta-testing" stage. This stage entails adapting the model to a sequence of incremental sessions, where each session introduces several novel classes with few-shot examples. Inspired by this, the model is meta-trained on base classes with the goal of mimicking the incremental learning scenario anticipated during the subsequent online incremental learning (i.e., evaluation). This ensures that the model is learned in a manner enabling effective adaptation to new classes with less forgetting.
\begin{figure*}[t]
    \centering
    \includegraphics[width=0.6\textwidth]{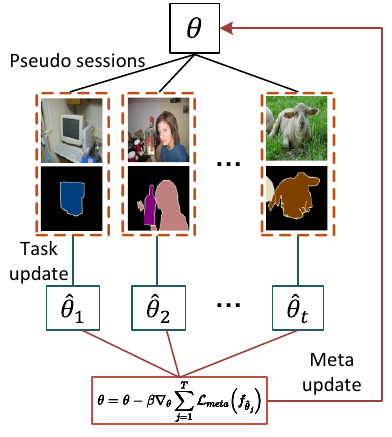}
    \caption{The meta-learning optimization strategy samples pseudo-sequential learning tasks on the base set to perform task training. The meta update process encourages the model to learn in a manner that preserves performance on old classes while effectively adapting to novel classes.}
    \label{figmeta}
\end{figure*}

\textbf{Sequential task sampling}. We replicate the evaluation process by utilizing the base classes. More precisely, we segregate the training images of base classes into distinct training and testing sets with no overlap. In each epoch, we initiate the training process by sampling a sequence of T tasks, $\mathcal{D}^s_{train/test}=\left\{\left(\mathcal{I}^j_{train/test}, \mathcal{M}^j_{train/test}\right)\right\}_{j=0}^T$, where $T$ is the actual incremental session number, and each session include training and testing image-mask pairs. We define $D^0$ as the pseudo base set, comprising more classes and training examples than subsequent tasks (e.g., $j >$ 0) in the few-shot setting. To mitigate the risk of the model overfitting to a particular sequence, we introduce random sampling of classes and their corresponding images
\begin{algorithm}\small
    \caption{Meta training process}
    \begin{algorithmic}[1]
        \REQUIRE $\theta ^ g,  \theta ^ {seg}$: pre-trained weights
        \REQUIRE $\mathcal{D}^0$: training set of base classes
        \WHILE {not converged}
        \STATE $\mathcal{D}^s_{train/test}=\left\{\left(\mathcal{I}^j_{train/test}, \mathcal{M}^j_{train/test}\right)\right\}_{j=0}^T$
        \STATE Initialize the models with $\mathcal{D}^0$.
        \STATE $\mathcal{D}_{meta} = \emptyset$
        \FOR{$j = 1,2,..., T$}
            \STATE \footnotesize$\mathcal{P} = Concat(\mathcal{P}_{old}, \mathcal{P}_{new})$
            \STATE  $\hat{\theta}^{g, seg}_j = \theta^{g, seg}-\alpha \nabla_{\theta^{g, seg}} \mathcal{L}_{CE}\left(\mathcal{I}_{train}^j, \mathcal{M}_{train}^j ; \theta \right)$
            \STATE  $\mathcal{D}_{meta} = \mathcal{D}_{meta} \cup \mathcal{D}^j_{test}$
            \STATE \footnotesize $\theta^{g,seg} = \theta^{g,seg} - \beta \nabla_{\theta^{g,seg}} \sum_{\left(\mathcal{I}, \mathcal{M}\right) \in \mathcal{D}_{meta}} \mathcal{L}_{meta}\left(\mathcal{I}, \mathcal{M}; \hat\theta^{g,seg}_j\right)$
        \ENDFOR
        \ENDWHILE
    \end{algorithmic}
\end{algorithm}

\textbf{Meta-training}. During the meta-training phase, for every sampled sequence $\mathcal{D}^s_{train/test}$, we introduce a prototype redistribution-oriented optimization approach grounded in Meta-Learning. We denote $\theta = \{\theta^f, \theta^g, \theta^{seg}\}$ as the parameter for the whole network, where $\theta^f, \theta^g, \theta^{seg}$ denote the parameters for backbone, prototype projection layer and segmentation head, respectively. We first conduct supervised training of $\theta$ on the base classes using cross-entropy loss ($\mathcal{L}_{CE}$). The meta-training procedure is illustrated in Alg. 1 and Fig.\ref{figmeta}. At the beginning of training on
each sequence, we define an empty cumulative meta test set $\mathcal{D}_{meta}$ to store the test images from previous tasks. At the $j^{th}$ task, we first generate the new class prototypes $\mathcal{P}_{new}$ and then concatenate it into the current prototype classifier $\mathcal{P}_{old}$. Subsequently, we start to perform fast adaptation to new classes and update $\theta^g$ and $\theta^{seg}$ via a few L gradient steps: 

\begin{equation}
\hat{\theta}^{g, seg}_j = \theta^{g, seg}-\alpha \nabla_{\theta^{g, seg}} \mathcal{L}_{CE}\left(\mathcal{I}_{train}^j, \mathcal{M}_{train}^j ; \theta \right),
\end{equation}
where $\mathcal{I}_{train}^j, \mathcal{M}_{train}^j$ are the images and labels for training $j^{th}$ pseudo task. The loss $\mathcal{L}_{C E}(,:)$ is computed on the output of the current model and the target label $\mathcal{M}_{train}^j$.

The adaptation process mimics the model's learning pattern for new classes during incremental sessions. Ideally, we aim for the adapted parameters to perform well in both the classes from the previous and current tasks. The meta-test set accumulated from previous tasks is used for evaluating how well the updated model resists catastrophic forgetting on old classes and adaptation on new classes. We append $\mathcal{D}^j_{test}$ to $\mathcal{D}_{meta}$ , and accordingly, the meta-objective is defined as:

\begin{equation}
\theta^{g,seg} = \theta^{g,seg} - \beta \nabla_{\theta^{g,seg}}\sum_{\left(\mathcal{I}, \mathcal{M}\right) \in \mathcal{D}_{meta}} \mathcal{L}_{meta}\left(\mathcal{I}, \mathcal{M}; \hat\theta^{g,seg}_j\right).
\end{equation}
Note that $\mathcal{L}_{meta}$ is a function designed to optimize $\theta^{g,seg}$ with the objective of achieving optimal performance through the redistribution of class prototypes as:
\begin{equation}
\mathcal{L}_{meta} = \mathcal{L}_{CE}(\mathcal{I}_{test},\mathcal{M}_{test} ) + \lambda \mathcal{L}_{r}. 
\end{equation}
When all N tasks are done, $\mathcal{D}_{meta}$ is reset to empty and we repeat the learning process from the random initialization and adaptation process.

In the online incremental learning stage, we execute Lines 5-7 of Alg. 1 to acquire knowledge about novel classes during evaluation. The steps outlined in Alg. 1 align with the evaluation protocol: after being trained on the current session, the model undergoes evaluation on all encountered classes so far. This meta-objective encourages our model to quickly adapt to novel classes without sacrificing remembering old ones.

\section{Experiments}

\subsection{Dataset}
We evaluate the proposed method on two widely used semantic segmentation datasets: PASCAL VOC 2012~\cite{VOC} and COCO~\cite{COCO}. Following established practices~\cite{protoiFSS}, we evenly partition the classes in PASCAL VOC and COCO into four folds, with each fold containing 5 and 20 categories, respectively. In the validation stage, three folds are used to form the base set, while the categories from the remaining fold are utilized for testing.

\subsection{Implementation details and evaluation metrics}
In all experiments, we employ ResNet-101~\cite{resnet} pre-trained on ImageNet as the feature extractor. Our configuration involves ASPP~\cite{deeplab} with a 1x1 convolutional layer as the segmentation head. We evaluate the performance of a method utilizing three mean intersection-over-union (mIoU) metrics: mIoU on base classes (mIoU-B), mIoU on new classes (mIoU-N), and the harmonic mean of the two (HM). Consistent with~\cite{protoiFSS}, all reported results are presented upon the completion of training in the final incremental session. Particularly, the single step means   while multi-step has multiple sessions: 5 sessions of 1 class on VOC and 4 sessions of 5 classes on COCO. 

\begin{table*}[t]
\caption{The experimental results on the PASCAL VOC 2012 dataset.}
\label{table 1}
\resizebox{\textwidth}{!}{
\begin{tabular}{l|cccccc|cccccc}
\toprule[1.5pt]
\multirow{3}{*}{Method} & \multicolumn{6}{c|}{Single step}                         & \multicolumn{6}{c}{Multi-step}                          \\ \cline{2-13} 
                        & \multicolumn{3}{c|}{1-shot} & \multicolumn{3}{c|}{5-shot} & \multicolumn{3}{c|}{1-shot} & \multicolumn{3}{c}{5-shot} \\ \cline{2-13} 
                        & mIoU-B    & mIoU-N   & HM     & mIoU-B    & mIoU-N    & HM     & mIoU-B    & mIoU-N   & HM     & mIoU-B    & mIoU-N   & HM     \\ \hline
Finetune                & 58.3    & 9.7     & 16.7   & 55.8    & 29.6     & 38.7   & 47.2    & 3.9     & 7.2    & 58.7    & 7.7     & 13.6   \\
WI~\cite{wi}                      & 62.7    & 15.5    & 24.9   & 64.9    & 21.7     & 32.5   & 66.6    & 16.1    & 25.9   & 66.6    & 21.9    & 33.0   \\
DWI~\cite{dwi}                    & 64.3    & 15.4    & 24.8   & 64.9    & 23.5     & 34.5   & \textbf{67.2}    & 16.3    & 26.2   &\textbf{67.6}    & 25.4    & 36.9   \\

MiB~\cite{mib}                    & 61.0    & 5.2     & 9.7    & \textbf{65.0}    & 28.1     & 39.3   & 43.9    & 2.6     & 4.9    & 60.9    & 5.8     & 10.5   \\
			SPN ~\cite{spn}                   & 59.8    & 16.3    & 25.6   &58.4    & 33.4
			     &42.5   & 49.8    & 8.1   & 13.9   & 61.6    & 16.3    & 25.8   \\
PIFS~\cite{protoiFSS}                   & 60.9    & 18.6    & 28.5   & 60.5    & 33.4     & 43.0   & 64.1    & 16.9    & 26.7   & 64.5    & 27.5    & 38.6   \\
HDMNet~\cite{HDMNet}                   & 57.7    & 16.4    & 25.5   & 58.1    & 34.9     & 43.6   & 52.2    & 15.6    & 19.0   & 55.0    & 14.7    & 23.2   \\
SRAA~\cite{advancing}                   & \textbf{65.2}    & 19.1    & 29.5   & 63.8    & \textbf{36.7}     & \textbf{46.6}   & 66.4    & 18.8    & 29.3   & 64.3    & 28.7    & 39.7  \\
Ours           & 63.4    & \textbf{19.7}    & \textbf{30.1}   & 61.6    & 35.8     & 45.3   & 65.5    & 20.4  & \textbf{31.1}   & 65.9    & \textbf{29.1}    & \textbf{40.4}  \\
\bottomrule[1.5pt]
\end{tabular}
}
\vspace{-1.0em}
\end{table*}

\begin{table*}[h]
\caption{The experimental results on the COCO dataset.}
\label{table 2}
\resizebox{\textwidth}{!}{
\begin{tabular}{l|cccccc|cccccc}
\toprule[1.5pt]
\multirow{3}{*}{Method} & \multicolumn{6}{c|}{Single step}                         & \multicolumn{6}{c}{Multi-step}                          \\ \cline{2-13} 
                        & \multicolumn{3}{c|}{1-shot} & \multicolumn{3}{c|}{5-shot} & \multicolumn{3}{c|}{1-shot} & \multicolumn{3}{c}{5-shot} \\ \cline{2-13}
                        & mIoU-B    & mIoU-N   & HM     & mIoU-B    & mIoU-N    & HM     & mIoU-B    & mIoU-N   & HM     & mIoU-B    & mIoU-N   & HM     \\ \hline
Finetune                & 41.2    & 4.1     & 7.5    & 41.6    & 12.3     & 19.0   & 38.5    & 4.8     & 8.5    & 39.5    & 11.5    & 17.8   \\
WI~\cite{wi}                     & 43.8    & 6.9     & 11.9   & 43.6    & 8.7      & 14.5   & \textbf{46.3}    & 8.3     & 14.1   & 46.3    & 10.3    & 16.9   \\
DWI~\cite{dwi}                    & \textbf{44.5}    & 7.5     & 12.8   & 44.9  & 12.1     & 19.1   & 46.2    & 9.2     & 15.3   & \textbf{46.6}    & 14.5    & 22.1   \\
MiB~\cite{mib}                   & 43.8    & 3.5     & 6.5    & 44.7    & 11.9     & 18.8   & 40.4    & 3.1     & 5.8    & 43.8    & 11.5    & 18.2   \\
SPN ~\cite{spn}             & 43.5    & 6.7     & 11.7   & 43.7    & 15.6     & 22.9   & 40.3    & 8.7   & 14.3   & 41.4    & 18.2 & 25.3   \\
PIFS~\cite{protoiFSS}            & 40.8    & 8.2     & 9.8   & 41.4    & 9.6     & 15.6   & 39.7    & 5.9    & 10.3   & 40.3    & 16.3    & 23.2   \\
HDMNet~\cite{HDMNet}                   & 39.5    & 5.6    & 9.8   & 40.1    & 13.6    & 20.3   & 39.7    & 6.5    & 11.2   & 41.4    & 12.6    & 19.3   \\
SRAA~\cite{advancing}                   & 41.2   & 9.3   & 15.2   & \textbf{46.2}    & 17.1    & 24.4   & 40.7    & 11.3    & 17.7  & 41.0    & 19.7  & 26.6  \\
Ours                    & 43.8    & \textbf{10.4}    & \textbf{16.7}   & 44.4    & \textbf{20.8}     & \textbf{28.3}   & 43.1    & \textbf{12.3}    & \textbf{19.1}   & 43.5    & \textbf{22.2}    & \textbf{29.4}   \\ 
\bottomrule[1.5pt]
\end{tabular}
}
\end{table*}
\subsection{Main results}
The outcomes of our method on the PASCAL VOC 2012 and COCO datasets are consolidated in Table~\ref{table 1} and Table~\ref{table 2}, respectively. We consider three baselines: Finetune, directly fine-tune the base model with new classes on each session; naive prototype classifier WI~\cite{wi} and its dynamic version DWI~\cite{dwi}; knowledge-distillation-based method MiB~\cite{mib}; FSS method HDMNet~\cite{HDMNet} and iFSS approach SRAA~\cite{advancing}. Our approach demonstrates superior performance in novel class adaptation across most settings for both PASCAL and COCO datasets. Additionally, it achieves state-of-the-art performance in terms of Harmonic Mean (HM) scores across all settings except for 5-shot single step, indicating that our approach effectively balances the retention of information about old classes while facilitating adaptation to new ones. Particularly noteworthy is our method's performance on the PASCAL dataset, where it achieves significantly higher novel class segmentation mIoU scores compared to all other methods, reaching 35.8\% and 29.1\% in single-step and multi-step settings, respectively. This surpasses the state-of-the-art method (SRAA) by 0.6\% and 1.8\% under 1-shot setting, respectively. Our meta-learning-based approach exhibits superior fast adaptation capability to novel classes without compromising base class segmentation accuracy, achieving competitive base class segmentation performance on both PASCAL and COCO datasets. In the single-step setting, all the new classes are introduced in a single session. When more samples are provided for a particular class, the model demonstrates improved adaptation to the novel class, as evidenced by the mIoU-N score, albeit with a potential decrease in performance for the base classes. This effect is mitigated in the multi-step setting, where our meta-learning approach effectively learns to resist forgetting through training across multiple sessions

On the COCO dataset, our approach showcases significantly greater improvements in HM scores compared to the state-of-the-art method SRAA~\cite{advancing}. For instance, in the task of 5-shot segmentation, our method's HM scores surpass those of SRAA by 3.9\% and 2.8\%, whereas the margins are only -1.3\% and 0.7\% on the PASCAL dataset. This highlights the effectiveness of our approach in tackling the more intricate challenges associated with a larger number of classes, which is particularly beneficial in real-world applications.
\begin{figure}[h]

               \centering

               \includegraphics[scale=0.65]{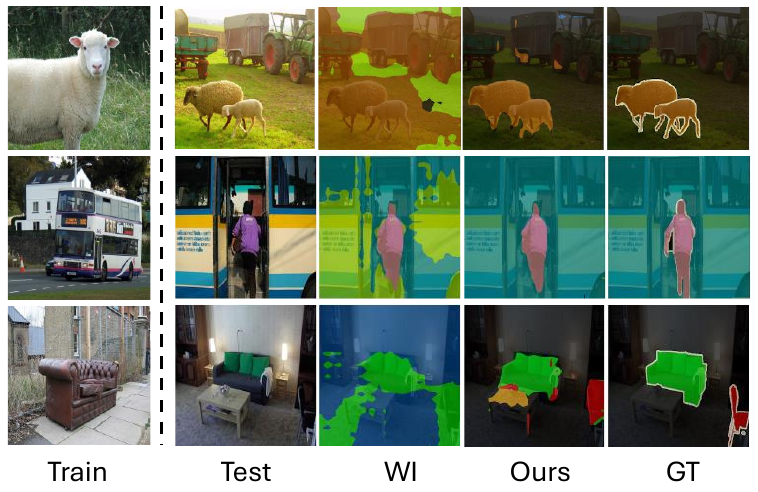}

               \caption{Visualization of multi-step results under shot setting on the PASCAL dataset.}
               \label{vis}
\vspace{-1.0em}
\end{figure}

In Fig. \ref{vis}, we showcase visualized segmentation results obtained from training under the multi-step incremental setup, using one training example for each novel class. In comparison to vanilla weight-printing (WI), which simply appends new class prototypes to the prototype classifier, our approach notably distinguishes novel classes like ``bus'' from the base class ``person'' and ``sheep'' from the background. Additionally, as observed in the third row, WI exhibits overfitting to the ``sofa'' and completely forgets the knowledge of the ``chair''. Our method, employing task-consistent meta-learning with prototype distribution loss, preserves the ability to segment learned classes while accurately adapting to new classes.

\subsection{Ablation study}





\begin{table}[h]
\vspace{0.5em} 
\centering
\caption{Ablation study of the meta-learning scheme and prototype redistribution loss on COCO, under the multi-step one-shot setting. $\mathcal{L}_{\text{inter}}=\sum_{i=1}^{N^b}\sum_{j=1}^{N^t} \text{Sim}(P^{t-1}_i, P^t_j)$ merely aims to minimize similarity between novel and base classes}
\label{table 3}
\setlength{\tabcolsep}{10pt} 
\scalebox{1.0}{
\begin{tabular}{cccc|ccc}
\toprule[1.5pt]
Baseline&Meta-learning & $\mathcal{L}_{\text{inter}}$ & $\mathcal{L}_{\text{r}}$ & Base & Novel & HM \\ \hline
       $\checkmark$& &  &       & \textbf{44.8} & 7.8 & 13.3 \\
$\checkmark$&$\checkmark$ & & & 42.5 & 10.6 & 17.0 \\
$\checkmark$&$\checkmark$ & $\checkmark$ & & 41.2 & 11.5 & 18.0 \\
$\checkmark$&$\checkmark$ & & $\checkmark$ & 43.1 & \textbf{12.3} & \textbf{19.1} \\
\bottomrule[1.5pt]
\end{tabular}
}
\vspace{0.5em} 
\end{table}

\textbf{Componet effectiveness.} As illustrated in the second row of Table~\ref{table 3}, introducing the meta-learning strategy, which trains the model in a manner aligned with the expected evaluation in the incremental sessions, significantly enhances novel class adaptation and mitigates catastrophic forgetting. The application of $\mathcal{L}_{inter}$ upon meta-learning results in a 0.9\% increase in novel class accuracy but induces a 1.3\% performance reduction in the base class. It suggests that merely focusing on minimizing the similarity between the new class and the old class prototypes while neglecting the drift of the base class can lead to prototype inconsistency before and after adaptation, resulting in knowledge forgetting.

\begin{table}[]
	\centering
		\caption{Ablations on backbones and prototype redistribution. ``fix" denotes that the backbone remains fixed during incremental steps, while ``update" means that the backbone continues to update. ``PR" indicates the addition of the prototype projection layer and the adoptation of the prototype redistribution loss $\mathcal{L}_r$.}
	\label{tab:backboneablation}
	\begin{tabular}{l|ccc}
		\toprule[1.5pt]
		Methods                & Novel & Base & HM   \\ \hline
		Backbone (fix)         & 7.2   & \textbf{44.1} & 12.4 \\
		Backbone (update)      & 7.8   & 36.0 & 12.8 \\
		Backbone (fix) + PR    & \textbf{10.6}  & 40.4 & \textbf{16.8} \\
		Backbone (update) + PR & 10.2  & 36.5 & 15.9 \\
		\bottomrule[1.5pt]
	\end{tabular}%
\end{table}

\textbf{Backbone and prototype redistribution.} To investigate the performance difference between frozen and updated backbones, we conduct comparison experiments using two baseline models. In these experiments, the pre-trained backbone is either kept fixed or updated during the incremental steps. The model with the fixed backbone is denoted as Backbone (fix), while the model with the updated backbone is referred to as Backbone (update). As shown in Table~\ref{tab:backboneablation}, Backbone (update) outperforms Backbone (fix) in terms of HM score, primarily due to its superior performance on novel classes. However, there is a significant drop in mIoU for base classes, indicating that updating the backbone without any constrain may lead to overfitting on new classes and result in catastrophic forgetting.

Then, we augment the model by appending a prototype projection layer after the backbone and applying prototype redistribution supervision to obtain the classifier. From the results of the last two rows of Table~\ref{tab:backboneablation}, the fixed version outperforms the updated counterpart by a significant margin in both novel and base class segmentation. This superiority is attributed to the fixed backbone's ability to retain information about the base classes, while ``PR" ensures that the prototypes in the subspace remain well-separated. These factors mitigate catastrophic forgetting and facilitate rapid adaptation.
\begin{figure}[h]

               \centering

               \includegraphics[scale=0.3]{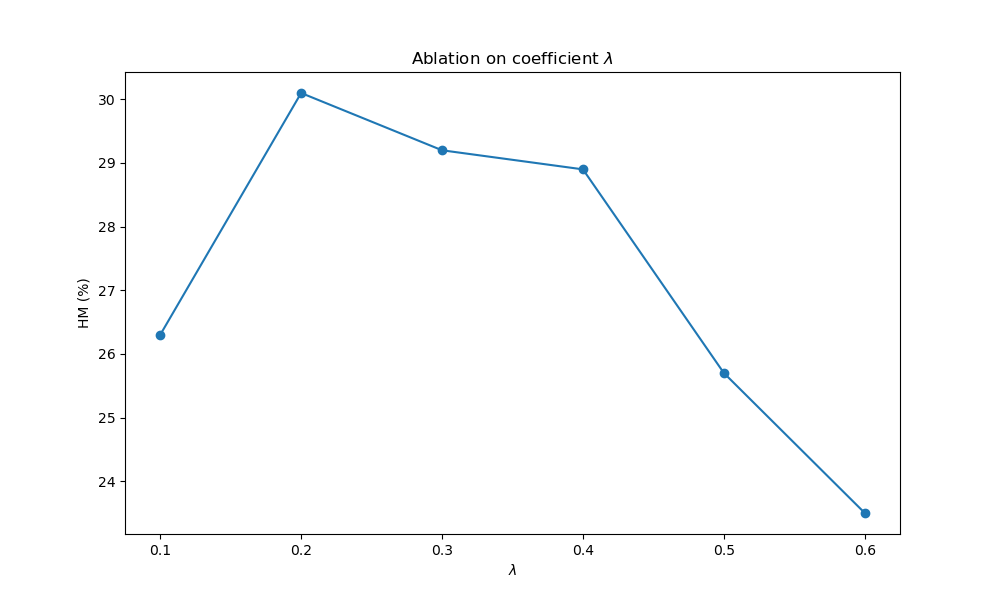}

               \caption{Ablation study on coefficient $\lambda$.HM performance in the Single step experiment under 1-shot setting.}
               \label{lambda}
\vspace{-1.0em}
\end{figure}

\textbf{Coefficient selection.} As shown in Fig. \ref{lambda}, we investigate the impact of the coefficient $\lambda$ in Equation 6 on the model's performance, testing lambda values from 0.1 to 0.6. The objective is to identify the optimal $\lambda$ that balances regularization with the model's ability to effectively learn new classes.

Our findings indicate that a $\lambda$ value of 0.3 achieves the best performance, as evidenced by the peak in the performance of HM under 1-shot setting. This optimal performance at lambda = 0.3 suggests an effective balance between forgetting and adaptability. Lower lambda values, closer to 0.1, may result in insufficient regularization, causing overfitting and poor generalization. Conversely, higher lambda values, approaching 0.6, could overly constrain the model, limiting its ability to adapt to novel classes and thereby degrading performance.

\section{Conclusion}
This work addresses a practical scenario of semantic segmentation that incrementally learns novel classes with a few examples. We propose a meta-learning-based approach, directly optimizing the network to acquire the ability to incrementally learn within the few-shot incremental setting. To alleviate catastrophic forgetting and overfitting problems, we introduce a prototype space re-distribution mechanism to dynamically update class prototypes during each incremental session. Extensive experiments on PASCAL and COCO benchmarks demonstrate that the proposed method facilitates a model learning paradigm for quick classes learning without forgetting.
%
%
%
\bibliographystyle{splncs04}
\bibliography{reference}
\end{document}